\pdfoutput=1

\documentclass[11pt]{article}

\usepackage[utf8]{inputenc}
\usepackage[T1]{fontenc}
\usepackage{lmodern}          
\usepackage{microtype}        

\usepackage[letterpaper,margin=1in]{geometry}
\usepackage{amsmath,amssymb,amsthm,mathtools}
\usepackage{bm}               

\usepackage{graphicx}
\graphicspath{{figures/}}     
\usepackage{booktabs}
\usepackage{multirow}
\usepackage{subcaption}       
\usepackage{xcolor}

\usepackage{parskip}
\usepackage{algorithm}
\usepackage{algpseudocode}

\usepackage[numbers,sort&compress]{natbib}

\usepackage{fancyhdr}
\fancypagestyle{plain}{\fancyhf{}\fancyfoot[C]{\thepage}}

\makeatletter
\renewenvironment{abstract}{%
  \small
  \begin{center}{\bfseries \abstractname\vspace{-.5em}\vspace{\z@}}\end{center}%
  \list{}{\listparindent\z@ \itemindent\z@ \rightmargin\leftmargin \parsep\parskip}%
  \item\relax
}{\endlist}
\makeatother

\usepackage{authblk}

\newcommand{\cindex}{C-index}

\usepackage[colorlinks=true,linkcolor=blue,citecolor=blue,urlcolor=blue]{hyperref}
\usepackage[capitalize,noabbrev]{cleveref}  
\crefname{algorithm}{Algorithm}{Algorithms}

\theoremstyle{definition}

\theoremstyle{remark}

\title{\textbf{A Multimodal Dataset for Survival Prediction in Resected Pancreatic Ductal Adenocarcinoma}}

\author[1, 2]{Anh-Tien Nguyen}
\author[2]{Mawuko Tettey}
\author[1]{Jacqueline Michelle Metsch}
\author[3]{Teresa Zimmer}
\author[3]{\authorcr Niklas Ullrich}
\author[3]{Mario Düker}
\author[3]{Sandra Rüngeling}
\author[3]{Kirsten Reuter-Jessen}
\author[3]{Tessa Rosenthal}
\author[4]{Lena-Christin Conradi}
\author[4]{Michael Ghadimi}
\author[5]{Alexander König}
\author[5]{Elisabeth Hessmann}
\author[5]{Volker Ellenrieder}
\author[3]{Philipp Ströbel}
\author[3]{Hanibal Bohnenberger\thanks{These authors jointly supervised this work.}\textsuperscript{,}\thanks{Corresponding authors: \texttt{hanibal.bohnenberger@med.uni-goettingen.de},\texttt{anne-christin.hauschild@uni-giessen.de}}}
\author[1]{Anne-Christin Hauschild\protect\footnotemark[1]\textsuperscript{,}\protect\footnotemark[2]}

\affil[1]{Institute for Predictive Deep Learning for Medicine and Healthcare, Giessen University, Germany}
\affil[2]{Department of Medical Informatics, University Medical Center Göttingen, Germany}
\affil[3]{Institute of Pathology, University Medical Center Göttingen, Germany}
\affil[4]{Department of General, Visceral and Pediatric Surgery, University Medical Center Göttingen, Germany}
\affil[5]{Department of Gastroenterology, Gastrointestinal Oncology and Endocrinology, University Medical Center Göttingen, Germany}

\date{}
\hypersetup{
  pdftitle={A Multimodal Dataset for Survival Prediction in Resected Pancreatic Ductal Adenocarcinoma},
  pdfauthor={First Author, Second Author, Third Author},
}

\begin{document}
\maketitle

\begin{abstract}

Survival research in pancreatic ductal adenocarcinoma (PDAC) is limited by the scarcity of datasets linking whole-slide histology with clinical, molecular, and long-term outcome data. We present a retrospective single-centre cohort of 302 patients who underwent PDAC resection at University Medical Center Göttingen. The dataset comprises 446 H\&E whole-slide images, clinicopathological variables, targeted sequencing data for 154 patients, and overall-survival outcomes. During follow-up, 253 patients died, and the median follow-up was 76 months.

To establish initial reference values, we evaluated fourteen survival-prediction configurations using identical five-repetition Monte Carlo cross-validation partitions. Ridge Cox regression using numeric clinicopathological variables achieved a mean concordance of $0.649 \pm 0.042$ and $0.652 \pm 0.046$ after adding \textit{KRAS} and \textit{TP53} mutation status. The image-only attention model achieved $0.603 \pm 0.030$, while multimodal fusion achieved $0.619 \pm 0.025$, the highest concordance among the neural models. These results establish promising initial benchmarks for future research using this pancreas-specific multimodal dataset, paving the way for external validation.
\end{abstract}

\vspace{0.5em}

\section{Introduction}
\label{sec:intro}

Pancreatic cancer remains one of the most lethal solid tumors, with an estimated 511,000 new cases and 467,000 deaths worldwide in 2022 alone \cite{bray2024global}. Although the five-year relative survival for cancer overall has now reached a historic milestone of 70\%, survival for pancreatic cancer remains below 15\%, the lowest of any major solid tumor \cite{siegel2026cancer}. This gap reflects both late-stage diagnosis and a limited ability to identify patients who have undergone resection into those likely to benefit from adjuvant therapy and those at highest risk of early recurrence. Reliable, individualized prognosis is therefore essential, helping guide treatment decisions, identify patients for clinical trials, and plan follow-up after surgery.

Clinical prognosis in resected pancreatic cancer still relies primarily on TNM stage together with a small set of pathology-report variables such as grade, margin status, and lymph node ratio. Yet multi-institutional validation of the current AJCC staging system has repeatedly shown wide survival overlap between stages and strong heterogeneity within the same stage \cite{allen2017multi}. Transcriptomic profiling has revealed that pancreatic tumors segregate into biologically distinct subtypes (e.g., classical versus basal-like, or squamous, immunogenic, and ADEX subtypes) with markedly different outcomes that are not captured by stage alone \cite{moffitt2015virtual,bailey2016genomic}. These findings indicate that stage-based, single-variable prognostic schemes disregard clinically relevant biological information, motivating approaches that jointly model histomorphology, clinical covariates, and molecular measurements rather than treating them as separate, competing sources of evidence.

Outside of pancreatic cancer, integrating whole-slide histology with molecular and clinical data has repeatedly improved survival prediction relative to any single modality. Early work combining convolutional features from H\&E images with genomic markers in glioma outperformed models built on either modality alone \cite{mobadersany2018predicting}. Other architectures such as Pathomic Fusion \cite{chen2020pathomic}, multimodal co-attention transformers \cite{chen2021multimodal}, and encoder-based multimodal survival models trained across cancer types \cite{valesilva2021long} have shown that attention- and fusion-based combinations of image and omics data improves model performance. In parallel, vision-language pathology
foundation models adapted through multi-granular prompt learning have improved label-efficient whole-slide representation learning under limited annotation \cite{nguyen2025mgpath}, pointing to a complementary route toward richer histology encoders. These results show that combining histology with molecular and clinical data is a promising direction in computational oncology.

Translating this to pancreatic cancer, however, is not easy. Pancreatic ductal adenocarcinoma is a particularly demanding setting for multimodal fusion as tumor cellularity in resection specimens is often low and diluted within a dense desmoplastic stroma \cite{erkan2012role}. Additionally, morphological presentation is heterogeneous even within a single tumor, and, as detailed in Section \ref{sec:related_datasets}, the public resources needed to jointly develop and validate such models (histology paired with both structured clinical follow-up and molecular data at survival-relevant scale) remain scarce. As a result, prior pancreas-specific studies have largely evaluated a single modality at a time, or have folded a small pancreatic subset into a pan-cancer multimodal benchmark rather than modeling the disease by itself (Section \ref{sec:related_cpath}).

In this work, we address this gap by introducing a large, multi-slide pancreatic cancer histopathology cohort with matched clinical variables and targeted mutation-panel sequencing (described in Section \ref{sec:related_datasets}), together with an initial multimodal benchmark for overall-survival prediction. Specifically, we (i) learn slide-level representations from H\&E whole-slide images (WSI) and combine them with structured clinical covariates and mutation status within a shared survival-modeling framework; (ii) benchmark this multimodal approach against unimodal and pairwise-modality baselines to quantify the contribution of histology, clinical data, and mutation status individually and in combination; and (iii) report these results as initial reference values rather than a definitive modeling strategy, so that the resulting dataset and benchmarks together provide a reproducible, pancreas-specific test set for future multimodal survival-prediction research. We hope that this resource lowers the barrier to future work integrating histomorphology and molecular profiling for pancreatic cancer, a disease where prognostic improvement remains most urgently needed.

\section{Related Work}
\label{sec:related_work}

\subsection{Public histopathology resources for pancreatic cancer}
\label{sec:related_datasets}

Progress in computational pathology for pancreatic cancer has been limited by
the scarcity of public whole-slide images (WSIs) linked to patient-level
outcomes. The public data landscape is dominated by two US consortium cohorts.
The Cancer Genome Atlas pancreatic adenocarcinoma cohort (TCGA-PAAD) contains
185 cases with multi-omic profiling and clinical follow-up
~\citep{raphael2017integrated}. After histologic review and quality control,
however, image-based studies typically retain only approximately 120-170
patients, often represented by a single diagnostic formalin-fixed slide
~\citep{saillard2023pacpaint,truntzer2024deep,vendittelli2024automatic}. The
Clinical Proteomic Tumor Analysis Consortium pancreatic cohort (CPTAC-PDA)
provides 557 WSIs from 168 subjects, together with extensive proteogenomic
characterization ~\citep{cao2021proteogenomic}. Thus, although
TCGA-PAAD and CPTAC-PDA provide the most comprehensive public resources for
outcome-oriented studies, each contains fewer than 200 patients.

Other pancreatic histopathology resources were created for narrower tasks and
do not provide longitudinal outcomes. PAIP~2021 contains 80 pancreatobiliary
WSIs with pixel-level annotations for perineural-invasion detection
~\citep{choi2021paip2021}, whereas PAIP~2023 provides 80 pancreatic H\&E image
patches for tumor-cellularity estimation ~\citep{paip2023challenge}. Both require
registration or a data-use agreement. Pancreatic tissue also occurs as a small
subset of pan-tissue collections for nuclei segmentation and mitosis detection
~\citep{gamper2019pannuke,mahbod2024nuinsseg,mahbod2021cryonuseg,aubreville2023comprehensive},
and several derived datasets add patch-level labels or annotations to
TCGA-PAAD without introducing independent patients
~\citep{komura2022universal,hou2020dataset,loeffler_2021_5320076}. Spatial-omics
resources pair pancreatic H\&E sections with transcriptomic measurements, but
currently comprise only tens of samples and lack survival endpoints
~\citep{jaume2024hest,elhossiny2026asynchronous}. Larger initiatives, including a
pancreatic research portal and a population-based virtual tissue repository,
remain accessible only by application
~\citep{oscanoa2025central,sanchez2024nci}. GTEx supplies normal-pancreas slides
that can serve as controls, but not matched pancreatic cancer outcomes
~\citep{sobin2025histologic}.

Consequently, the available data are fragmented across diagnostic,
segmentation, molecular, and spatial-omics tasks. To our knowledge, TCGA-PAAD
and CPTAC-PDA remain the only public cohorts that combine pancreatic cancer
WSIs with both molecular profiles and survival information, and neither offers
a large, independent population for survival-model development and external
validation.

\subsection{Computational pathology of pancreatic cancer}
\label{sec:related_cpath}

Existing methods can be grouped into diagnosis, molecular characterization,
prognosis, and tissue quantification. For diagnosis, convolutional and
multiple-instance learning (MIL) models have detected pancreatic ductal
adenocarcinoma (PDAC) in resection WSIs and endoscopic-ultrasound-guided
fine-needle biopsies, while related systems support rapid on-site cytologic
evaluation ~\citep{fu2021automatic,naito2021deep,zhang2022deep}. Strong internal
performance does not necessarily translate across cohorts. For example, a
model trained on TCGA and GTEx achieved an internal AUC of 0.997 but only 0.668
when evaluated on CPTAC; the reciprocal CPTAC-to-TCGA/GTEx transfer achieved an
AUC of 0.839 ~\citep{carrillo2023performance}. The same study showed that
dataset-specific features could dominate class-specific morphology,
illustrating the need for independent, geographically distinct validation
data.

Histologic morphology also contains information about tumor biology.
Pan-cancer and pancreas-specific studies have predicted genomic alterations
and transcriptomic subtypes directly from H\&E images
~\citep{kather2020pan,arslan2024systematic,ahmadvand2024deep}. PACpAInt, for
example, used 1{,}796 slides from 598 patients to predict basal-like and
classical PDAC subtypes and map intratumoral heterogeneity; only its
126-patient TCGA validation cohort is publicly available
~\citep{saillard2023pacpaint}. This distinction is important because large
multi-slide studies can characterize heterogeneity more effectively, yet their
institutional data generally cannot be redistributed.

For prognosis, prior work has extracted interpretable features such as stromal
and lymphocytic composition ~\citep{tan2025stroma}, tumor--stroma ratio
~\citep{vendittelli2024automatic}, and spatial heterogeneity
~\citep{shao2026ai}. Other approaches learn prognostic representations using
attention-based MIL or transcriptome-linked morphologic clusters
~\citep{truntzer2024deep,hu2024construction,takamatsu2026prognostic}. Pan-cancer
multimodal models have also included TCGA-PAAD, although reported concordance
indices are approximately 0.6 ~\citep{chen2022pan}. In parallel, segmentation
models quantify tumor, stroma, and residual disease after neoadjuvant therapy,
including an international validation study involving 528 patients and four
scanner types
~\citep{vendittelli2022automatic,janssen2021artificial,janssen2024artificial}.
Collectively, these studies demonstrate the prognostic value of pancreatic
histomorphology, but most larger development and validation cohorts remain
institutional or consortium resources that are not publicly redistributable.

General-purpose pathology foundation models have broadened the range of
transferable representations, but pancreas-specific evaluation remains
limited. UNI ~\citep{chen2024towards}, CONCH ~\citep{lu2024visual}, and
Prov-GigaPath ~\citep{xu2024whole} did not report a dedicated pancreatic
benchmark in their original evaluations. Virchow included pancreatic tissue
within a pooled pan-cancer detection task ~\citep{vorontsov2024foundation}, and
CHIEF used pancreatic datasets for cancer detection but not for a dedicated
pancreatic survival analysis ~\citep{wang2024pathology}. More recently, PRISM2
included TCGA-PAAD among 13 TCGA disease-specific survival tasks
~\citep{vorontsov2026end}. This result demonstrates the transferability of
slide-level foundation-model representations, but PAAD remains one component
of a pan-cancer benchmark rather than an independent, pancreas-specific
survival validation cohort.

Against this background, we introduce an institutional multimodal PDAC dataset collected at University Medical Center Göttingen. The dataset combines multiple H\&E WSIs per patient with structured clinicopathological variables, targeted panel sequencing, and long-term overall-survival data. It complements TCGA-PAAD and CPTAC-PDA by providing an independent patient population with linked histological, clinical, molecular, and outcome information. To characterise the dataset and facilitate future comparisons, we report initial survival-prediction benchmarks using individual modalities and multimodal combinations. These experiments are intended to establish reference performance for this cohort rather than to propose a definitive survival-prediction model. The dataset therefore provides a resource for future research on cross-cohort validation, patient-level multi-slide aggregation, multimodal learning, and pancreas-specific evaluation of pathology foundation models.

\section{Dataset}
\label{sec:dataset}

\subsection{Study population and data sources}

We conducted a retrospective, single-centre cohort study of patients who underwent surgical resection for pancreatic ductal adenocarcinoma (PDAC) at University Medical Center Göttingen between 2000 and 2021. Patients who had received neoadjuvant treatment were excluded. Patients were eligible if they had documented vital status and overall survival (OS) time, an OS exceeding two months, and at least one available archival haematoxylin and eosin (H\&E)-stained whole-slide image. The two-month threshold was applied to reduce the influence of early postoperative
mortality on the survival benchmark. Of the 631 patients screened, 302 met all eligibility criteria and comprised the final study cohort. For each case, the H\&E-stained slide or slides with the highest tumour cell content were selected and digitised using a Philips scanner at $40\times$ magnification. During follow-up, 253 patients (83.8\%) died and 49 (16.2\%) were alive at last follow-up and were therefore censored.

Kaplan-Meier analysis of the analysis cohort yielded a median OS of 21.0
months, with estimated survival probabilities of 69.5\%, 42.9\%, 29.6\%, and
15.5\% at 1, 2, 3, and 5 years, respectively. Median follow-up, estimated by
the reverse Kaplan--Meier method, was 76.0 months. Cohort characteristics are
summarised in Table~\ref{tab:cohort}.

\begin{table}[H]
\centering
\caption{Characteristics of the WSI analysis cohort ($n=302$ patients).}
\label{tab:cohort}
\small
\setlength{\tabcolsep}{3pt}
\begin{tabular}{@{}llc@{}}
\toprule
Characteristic & Value & Missing, $n$ (\%) \\
\midrule
Age at diagnosis, years - median [IQR] & 68 [62-74] & 0 (0.0) \\
Sex, male - $n$ (\%) & 179 (59.3) & 0 (0.0) \\
Histologic grade - $n$ (\%) &  & 0 (0.0) \\
\quad G1 & 7 (2.3) &  \\
\quad G2 & 206 (68.2) &  \\
\quad G3 & 89 (29.5) &  \\
Pathological T category - $n$ (\%) &  & 2 (0.7) \\
\quad pT1 & 19 (6.3) &  \\
\quad pT2 & 79 (26.2) &  \\
\quad pT3 & 190 (62.9) &  \\
\quad pT4 & 12 (4.0) &  \\
Pathological N category - $n$ (\%) &  & 0 (0.0) \\
\quad pN0 & 76 (25.2) &  \\
\quad pN1 & 190 (62.9) &  \\
\quad pN2 & 36 (11.9) &  \\
Resection margin R1/R2 - $n$ (\%) & 100 (33.1) & 5 (1.7) \\
Lymph node ratio - median [IQR] & 0.13 [0.00-0.29] & 1 (0.3) \\
Examined regional lymph nodes - median [IQR] & 16 [10-21] & 1 (0.3) \\
UICC stage group - $n$ (\%) &  & 2 (0.7) \\
\quad I & 31 (10.3) &  \\
\quad II & 210 (69.5) &  \\
\quad III & 46 (15.2) &  \\
\quad IV & 13 (4.3) &  \\
Tumour sequenced - $n$ (\%) & 154 (51.0) &  \\
\quad \textit{KRAS} pathogenic - $n$ (\% seq.) & 133 (86.4) &  \\
\quad \textit{TP53} pathogenic - $n$ (\% seq.) & 78 (50.6) &  \\
Whole-slide images - $n$ & 446 &  \\
Slides per patient - median [IQR] & 1 [1--2] &  \\
Deaths during follow-up - $n$ (\%) & 253 (83.8) &  \\
Median OS (Kaplan-Meier), months & 21.0 &  \\
1-/2-/3-/5-year OS, \% & 69.5/42.9/29.6/15.5 &  \\
Median follow-up (reverse KM), months & 76.0 &  \\
\bottomrule
\end{tabular}
\end{table}

\subsection{Clinical variables}

The primary feature set comprised seven routinely available clinicopathological variables: age at diagnosis (years), sex, histological grade (G1-G3), pathological T category (pT1-pT4), pathological N category (pN0-pN2), resection-margin status (R0 vs.\ R1/R2), and lymph node ratio (LNR). The pT and pN categories were taken from the original pathology reports. LNR was defined as the number of tumour-involved regional lymph nodes divided by the total number of examined regional lymph nodes. LNR was selected instead of the composite UICC stage group because stage largely reflects pT and pN in resected PDAC, whereas LNR captures variation in relative nodal involvement within pN categories. An alternative feature set replacing LNR with UICC stage group (I-IV) was evaluated in a sensitivity analysis. Among patients with available nodal counts, data-quality checks confirmed that those classified as pN0 had no tumour-involved nodes and that the number of involved nodes never exceeded the number examined.

Missing covariate data were uncommon: pT was missing in 2 patients (0.7\%), resection-margin status in 5 (1.7\%), LNR in 1 (0.3\%), and UICC stage in 2 (0.7\%). Age, sex, histological grade, and pN were complete. Missing entries were imputed using medians calculated exclusively from the corresponding training split.

\paragraph{Targeted genomic profiling.}
Targeted genomic profiling was performed on tumour DNA using a custom-designed next-generation sequencing (NGS) panel covering 102 cancer-associated genes which are listed in the Appendix ~\ref{fig:gene-availability} and ~\ref{fig:mutations}. The panel targeted all coding exons for the majority of genes.

Tumour DNA was extracted from FFPE tissue sections of the resection specimen. Tumour-rich regions were identified on H\&E-stained sections, and macrodissection was performed to enrich for tumour content. Only samples with an estimated minimum tumour cell content of 60\% were processed further. Genomic DNA was isolated using the automated Maxwell® system in combination with the CSC FFPE DNA Purification Kit (Promega). DNA was fragmented on a Covaris ME220 instrument to an average fragment size of 150–200 bp and quantified using the Qubit™ dsDNA High Sensitivity Assay Kit (Thermo Fisher Scientific) according to the manufacturer's instructions.

Library preparation was carried out using the SureSelect XT HS2 Kit (Agilent) with an input of 100 ng fragmented DNA, including end repair, A-tailing, adapter ligation incorporating unique molecular identifiers (UMIs) and PCR amplification. Target enrichment was performed by in-solution hybrid capture using the custom gene panel. Library preparation was automated on the NGS Star system (Hamilton). Equimolar amounts of libraries were pooled and sequenced with 150-bp paired-end reads on an Illumina NextSeq 500 platform.

Raw sequencing data (FASTQ files) were generated on-instrument and processed using CLC Genomics Workbench (version 21.05, Qiagen) for read alignment, variant calling and downstream analysis. Reads were aligned to the human reference genome GRCh37/hg19. Samples with a mean sequencing coverage below 100× were considered failed and excluded. Variant calling included single-nucleotide variants and small insertions or deletions. Variants were retained if they fulfilled predefined quality criteria, including a minimum read count greater than 50, minimum coverage at the variant position greater than 60×, a variant allele frequency greater than 5\%, and support by at least ten unique start and end sites. UMIs were used for read annotation and duplicate collapsing using the standard UMI-processing workflow implemented in CLC Genomics Workbench.

Detected variants were interpreted using established clinical knowledge bases (ClinVar and OncoKB) and classified as pathogenic or likely pathogenic, variants of uncertain significance, or benign or likely benign. For the present analysis, KRAS and TP53 mutation status was defined as the presence of at least one pathogenic or likely pathogenic variant in the respective gene.

\subsection{Outcome encoding for individual survival distributions}

The prediction task was to estimate an individual survival distribution (ISD) on a discrete time grid comprising eleven intervals: ten consecutive six-month intervals spanning $[0,60)$ months and a final open-ended interval $[60,\infty)$. For each patient $i$, the observed time $t_i$ (time to death or censoring, in months) was mapped to a zero-indexed time bin,
$y_i=\min\!\big(\lfloor t_i/6 \rfloor,\,10\big)$
and paired with an event indicator $\delta_i$, where $\delta_i=1$ denoted death and $\delta_i=0$ denoted right censoring. The pair $(y_i,\delta_i)$ constituted the observed outcome label. For censored patients, this label indicated survival up to the censoring time rather than death within the assigned interval. The ISD, represented by the predicted survival function $\hat S_i(t)$, was a model output evaluated against these observed outcomes, not a directly observed ground-truth distribution.

\subsection{Cross-validation design}
\label{sec:splits}

Model performance was evaluated using five repetitions of Monte Carlo cross-validation(MCCV). In each repetition, the full cohort of 302 patients was randomly partitioned at the patient level into mutually exclusive training ($n=152$), validation ($n=75$), and test ($n=75$) sets in an approximately 2:1:1 ratio, with stratification by event status. Partitions were sampled independently across repetitions. The five partitions were generated once using a fixed random seed and stored for reuse, ensuring that all models and feature representations were trained and evaluated on identical patient splits. Although the three sets were disjoint within each repetition, test sets could overlap across repetitions, as expected in Monte Carlo cross-validation.

\section{Benchmark}
\label{sec:benchmark}

\subsection{Overview}
 
We evaluated OS prediction using three information sources: routine clinicopathological variables, tumour mutation status, and diagnostic whole-slide images (WSIs). The benchmark distinguished three design components: the \emph{input configuration} (which information sources were included), the \emph{encoding backbone} (how inputs were transformed into features), and the \emph{survival model} (how these features were mapped to survival predictions). Fourteen configurations were evaluated (Table~\ref{tab:benchmark}), encompassing numeric covariates, text representations of the same covariates generated using two frozen language encoders, WSI-derived features, and multimodal image-text fusion.

Model performance was evaluated using five repetitions of Monte Carlo cross-validation. In each repetition, patients were randomly assigned to training ($n=152$), validation ($n=75$), and test ($n=75$) sets in an approximately 2:1:1 ratio, with stratification by event status. Patient-level partitions were generated independently across repetitions and reused for all models. Evaluation was performed on the same held-out test sets using a common outcome definition and performance metric. Two baseline models were included: ridge-regularised Cox regression and a deep survival network, both using the same numeric covariates. These baselines were used to compare survival modelling approaches and assess whether alternative input representations or additional data sources improved predictive performance.

\subsection{Input representations}

\paragraph{Numeric covariates.}
Seven routinely available clinicopathological variables including age at diagnosis, sex, histological grade, pT, pN, resection-margin status, and lymph node ratio were encoded as a seven-dimensional numeric vector. The extended feature set included two additional binary indicators for pathogenic mutations in \textit{TP53} and \textit{KRAS}. Mutation status was recorded as missing for patients without sequencing data. Within each cross-validation repetition, missing values were imputed using per-variable training-set medians, and all variables were standardised using training-set means and standard deviations. These preprocessing parameters were applied unchanged to the validation and test sets.

\paragraph{Clinical text.}
Each patient's covariates were converted into a short description using fixed sentence templates and standard TNM terminology. Missing values were represented as ``not reported'' rather than imputed. The descriptions therefore used the same source variables as the numeric inputs, while retaining missingness explicitly.

Descriptions were encoded using two frozen models: (i) BioClinical ModernBERT~\citep{sounack2025bioclinical}, a long-context encoder based on the ModernBERT architecture~\citep{warner2025smarter} and pretrained on biomedical and clinical text; and (ii) the text encoder of CONCH~\citep{lu2024visual}, a vision-language foundation model for computational pathology that produces 512-dimensional text embeddings.



\paragraph{Whole-slide images.}
WSIs were divided into non-overlapping patches. Each patch was encoded using the frozen CONCH image encoder~\citep{lu2024visual} to obtain a 512-dimensional embedding. CONCH's image and text encoders produce representations in a shared embedding space, which was used for image-text fusion. 


\subsection{Whole-slide token selection}
\label{sec:focus}

The patch set of each slide was reduced using the deterministic, concept-guided FOCUS procedure~\citep{guo2025focus}. Patch relevance was assessed using contrast scoring based on similarity to a bank of 22 pancreatic-pathology text prompts (14 tumour-related and 8 background concepts), embedded with the frozen CONCH text encoder. Redundant neighbouring patches were pruned using an embedding-similarity threshold of 0.9.


\subsection{Survival models}

All models produced survival estimates on a common evaluation time grid. These estimates were converted to scalar risk scores using the same definition for concordance evaluation.

\paragraph{Ridge Cox model.}
A Cox proportional-hazards model~\citep{cox1972regression} with $\ell_2$ (ridge) regularisation was fitted using \texttt{scikit-survival}~\citep{polsterl2020scikit}.

\paragraph{Individual Survival Distribution (ISD).}
Each neural model comprised a modality-specific encoder followed by a discrete-time survival head with the same architecture across models. The head mapped the encoded features to eleven logits, one per time interval. The logits were transformed into interval-specific hazards using the sigmoid function,

$$
h_i(k)=\sigma(z_{ik}), \qquad k=0,\ldots,10.
$$

The individual survival distribution (ISD)~\citep{haider2020effective} was represented by the discrete survival function

$$
\hat S_i(k)=\prod_{j=0}^{k}\bigl[1-h_i(j)\bigr],
$$

with $\hat S_i(-1)=1$. Training minimised the negative log-likelihood for the encoded right-censored outcomes~\citep{zadeh2020bias,chen2022pan}:
\begin{equation}
\mathcal{L}
= -\sum_{i:,\delta_i=1}
\Bigl[\log \hat S_i(y_i-1)+\log h_i(y_i)\Bigr]
-\sum_{i:,\delta_i=0}\log \hat S_i(y_i),
\label{eq:nll}
\end{equation}
where $y_i$ denotes the observed time-bin index and $\delta_i$ indicates death ($1$) or right censoring ($0$).


\paragraph{Attention-based WSI model.}
Histological information was modelled using the concept-guided FOCUS architecture~\citep{guo2025focus}. The selected patch tokens were projected into the model's concept space and processed using multi-head cross-attention. Tumour-concept embeddings served as queries, while the projected patch tokens served as keys and values. The resulting concept-specific representations were combined using learned attention pooling and passed to the discrete-time survival head described above.


\paragraph{Multimodal co-attention fusion.}
Multimodal fusion was implemented using the co-attention mechanism of MCAT~\citep{chen2021multimodal}. The genomic tokens used in the original architecture were replaced by nine CONCH-derived text tokens including one for each of the seven clinicopathological variables and two mutation indicators. Because the image and text tokens were generated by the paired CONCH encoders, they occupied a shared pretrained embedding space before fusion. Each text token served as a query in single-head scaled dot-product attention over the patient's WSI tokens, which served as keys and values. This produced one image-conditioned representation for each input variable.

The image-conditioned representations and the original text tokens were aggregated separately using gated attention pooling~\citep{ilse2018attention}. The two pooled vectors were concatenated, projected to a 64-dimensional patient representation, and passed to the discrete-time survival head described above. Only the co-attention component of MCAT was adopted; the transformer encoders from the original architecture were not used.

\subsection{Evaluation and statistical analysis}
 
The primary performance measure was Harrell's concordance index~\citep{harrell1982evaluating}, which assesses the ability of predicted risk scores to rank survival outcomes using patient pairs comparable under right censoring. For each model, concordance was calculated on the held-out test set of each repetition and summarised as mean $\pm$ standard deviation across the five Monte Carlo repetitions. All models additionally generated patient-specific ISDs, represented by predicted survival probabilities at the prespecified six-month evaluation time points.

Continuous variables were summarised as medians with interquartile ranges (IQRs), and categorical variables as counts and percentages. Survival curves and median follow-up were estimated using the Kaplan--Meier and reverse Kaplan-Meier methods, respectively.

 \section{Results}

Table~\ref{tab:benchmark} reports test-set concordance for all fourteen evaluated configurations. These analyses were conducted to establish initial reference values for survival prediction in the institutional PDAC cohort. They are intended to support future model development and comparison using this dataset, rather than to identify a definitive modelling strategy. Five observations summarise the benchmark results.

\paragraph{Numeric clinicopathological variables provided the primary reference baselines.}
Using the seven numeric clinicopathological variables, ridge-regularised Cox regression achieved a mean \cindex{} of $0.649 \pm 0.042$, whereas the deep discrete-time model achieved $0.599 \pm 0.040$. These results provide linear and neural reference values for the clinical feature set. The difference between the models should be interpreted in light of their fitting protocols: the Cox model was fitted using the combined training and validation sets ($n=227$), whereas the neural model was fitted on the training set ($n=152$), with the validation set used for epoch selection. The comparison therefore does not isolate the effect of model class alone.

\paragraph{Adding mutation status produced little change in the benchmark results.}
After the mutation status of \textit{TP53} and \textit{KRAS} was added, the mean \cindex{} of the Cox model changed from $0.649 \pm 0.042$ to $0.652 \pm 0.046$. The latter was the highest mean concordance observed in the benchmark, although the numerical increase was only 0.003. For the deep discrete-time model, concordance changed from $0.599 \pm 0.040$ to $0.589 \pm 0.060$. Thus, the addition of mutation status did not produce a consistent improvement across the two modelling approaches. These values serve as initial reference results and do not establish the incremental prognostic value of the mutation variables.

\paragraph{Frozen text embeddings provided alternative baselines for the same covariates.}
For the mutation-extended feature set, Cox regression achieved a mean \cindex{} of 0.652 using the numeric variables, compared with 0.594 using BioClinical ModernBERT embeddings and 0.538 using CONCH text embeddings. For the clinical-only feature set, performance also varied according to the combination of text encoder and survival model. CONCH embeddings achieved mean concordance values of 0.594 with the deep reader and 0.562 with the Cox reader, whereas BioClinical ModernBERT embeddings achieved 0.558 and 0.594, respectively. These results establish baseline performance for frozen text representations of the tabular variables. Differences between encoders or readers are reported descriptively because the benchmark was not designed to determine their underlying causes.

\paragraph{Whole-slide images supported survival prediction without clinical covariates.}
Using WSIs alone, the concept-guided attention model achieved a mean \cindex{} of $0.603 \pm 0.030$, the highest value among the single-modality neural configurations. Ridge Cox regression applied to the mean-pooled image representation achieved $0.570 \pm 0.021$. Although both configurations used CONCH patch embeddings and identical data partitions, they differed in token selection, aggregation, survival model, and fitting protocol. The difference of 0.033 therefore describes the performance of the complete image-analysis pipelines and cannot be attributed to aggregation alone.

\paragraph{Multimodal fusion provided the highest neural benchmark.}
Co-attention fusion of WSI tokens with CONCH-derived tokens representing the clinical and mutation variables achieved a mean \cindex{} of $0.619 \pm 0.025$. This was the highest mean concordance among the neural configurations and represented numerical increases of 0.016 over the image-only attention model, 0.071 over the corresponding text-only model, and 0.030 over the deep model using the mutation-extended numeric feature set. For the linear Cox models, the combined pooled image-text representation achieved a mean \cindex{} of 0.594, compared with 0.570 for the image representation and 0.538 for the text representation.

The fused neural model remained below the clinical-only and mutation-extended numeric Cox baselines ($0.619$ vs.\ $0.649$ and $0.652$, respectively). Its across-repetition standard deviation was $0.025$, compared with $0.042$ for the clinical-only Cox model; given the five repetitions, this difference is descriptive rather than evidence of greater stability. The fusion model additionally generated per-variable co-attention maps over the WSI tokens. These maps provide material for subsequent analysis but are not treated as validated explanations in the present benchmark.

Overall, these results establish initial performance references for clinical, molecular, histological, and multimodal survival models applied to the institutional cohort. Further model development and external validation will be required to assess whether these findings generalise beyond this dataset.

\begin{table}[H]
\centering
\caption{Test-set concordance (mean $\pm$ SD over five Monte Carlo
repetitions). Clinical feature set: age, sex, grade, pT, pN, margin status, LNR. Mutation feature set: TP53, KRAS.}
\label{tab:benchmark}
\small
\begin{tabular}{llcc}
\toprule
Input representation & Backbone & Model & C-index \\
\midrule
Clinical & Numeric & Cox & $0.6489 \pm 0.0424$ \\
Clinical & Numeric & ISD & $0.5992 \pm 0.0397$ \\
Clinical & CONCH & Cox & $0.5623 \pm 0.0363$ \\
Clinical & CONCH & ISD & $0.5935 \pm 0.0105$ \\
Clinical + Mutation & Numeric & Cox & $\mathbf{0.6516 \pm 0.0457}$ \\
Clinical + Mutation & Numeric & ISD & $0.5893 \pm 0.0600$ \\
Clinical + Mutation & BioClinical ModernBERT & Cox & $0.5942 \pm 0.0456$ \\
Clinical + Mutation & BioClinical ModernBERT & ISD & $0.5578 \pm 0.0238$ \\
Clinical + Mutation & CONCH & Cox & $0.5377 \pm 0.0545$ \\
Clinical + Mutation & CONCH & ISD & $0.5476 \pm 0.0428$ \\
WSI & CONCH & Cox & $0.5699 \pm 0.0209$ \\
WSI & CONCH & ISD & $0.6025 \pm 0.0302$ \\
WSI + Clinical + Mutation & CONCH & Cox & $0.5944 \pm 0.0181$ \\
WSI + Clinical + Mutation & CONCH & ISD & $0.6189 \pm 0.0245$ \\
\bottomrule
\end{tabular}
\end{table}

\section{Conclusion}

We present a multimodal dataset from an institutional cohort of patients with resected pancreatic ductal adenocarcinoma, linking digitised H\&E whole-slide images with routinely collected clinicopathological variables, targeted mutation data, and long-term overall-survival outcomes. By integrating these data at the patient level, the dataset expands the limited resources available for pancreas-specific computational pathology and enables clinical, molecular, histological, and multimodal approaches to be evaluated using a common outcome definition.

To provide initial reference values, we evaluated fourteen combinations of input representation, feature-encoding backbone, and survival model. Ridge-regularised Cox regression using numeric clinicopathological and mutation variables achieved the highest mean concordance in this benchmark (\cindex{} $0.652 \pm 0.046$). The image-only attention model supported survival discrimination from histology alone (\cindex{} $0.603 \pm 0.030$), while multimodal co-attention achieved the highest concordance among the neural configurations (\cindex{} $0.619 \pm 0.025$). Representing tabular covariates using frozen text embeddings did not improve mean concordance over their numeric representation. These findings should be interpreted as initial dataset benchmarks rather than evidence of an optimal survival-modelling strategy.

The dataset and its accompanying benchmarks provide a foundation for future research on pancreas-specific foundation models, patient-level WSI aggregation, multimodal survival modelling, and cross-cohort validation.

\section*{Acknowledgements}
This work is supported in part by funds from the German Ministry of Education and Research (BMBF) under grant agreements \textit{No. 01D2208A} and \textit{No. 01KD2414A} (project FAIrPaCT). The authors gratefully acknowledge the computing time granted by the KISSKI project. The calculations for this research were conducted with computing resources under the project \textit{kisski-umg-fairpact-2}.  The authors also acknowledge the computing time granted by the Resource Allocation Board and provided on the supercomputer Emmy/Grete at NHR-Nord@Göttingen as part of the NHR infrastructure. The calculations for this research were conducted with computing resources under the project \textit{nim00014}. Anh-Tien Nguyen was a member of the Ph.D. program "Genome Science" - International Max Planck Research School. \\
We gratefully acknowledge support from the hessian.AI Service Center (funded by the Federal Ministry of Research, Technology and Space, BMFTR, grant no. 16IS22091) and the hessian.AI Innovation Lab (funded by the Hessian Ministry for Digital Strategy and Innovation, grant no. S-DIW04/0013/003). This work is supported in part by funds from CancerScout project (German Federal Ministry of Education and Research, BMBF, grant no. 13GW0451A).

\section{Data availability statement}
The datasets generated and/or analysed during the current study are available from the corresponding author on reasonable request.

\section{Ethics statement}
The study was approved by the Ethics Committee of University Medical Center Göttingen (no. 24-4-20).

\bibliographystyle{unsrtnat}   
\bibliography{references}

\newpage
\appendix
\section{Appendix}

\section{Mutation Profiles of Sequenced Patients}

\subsection{Targeted sequencing data availability}
Among the 302 patients with whole-slide image features, targeted sequencing results were available for 154 (51.0\%) and unavailable for 148 (49.0\%). The availability matrix shows that all 102 displayed genes were assessed in each sequenced patient. Missingness therefore occurred at the patient level, with no gene-level gaps among patients with panel results.

\begin{figure}[p]
    \centering
    \includegraphics[width=\textwidth]{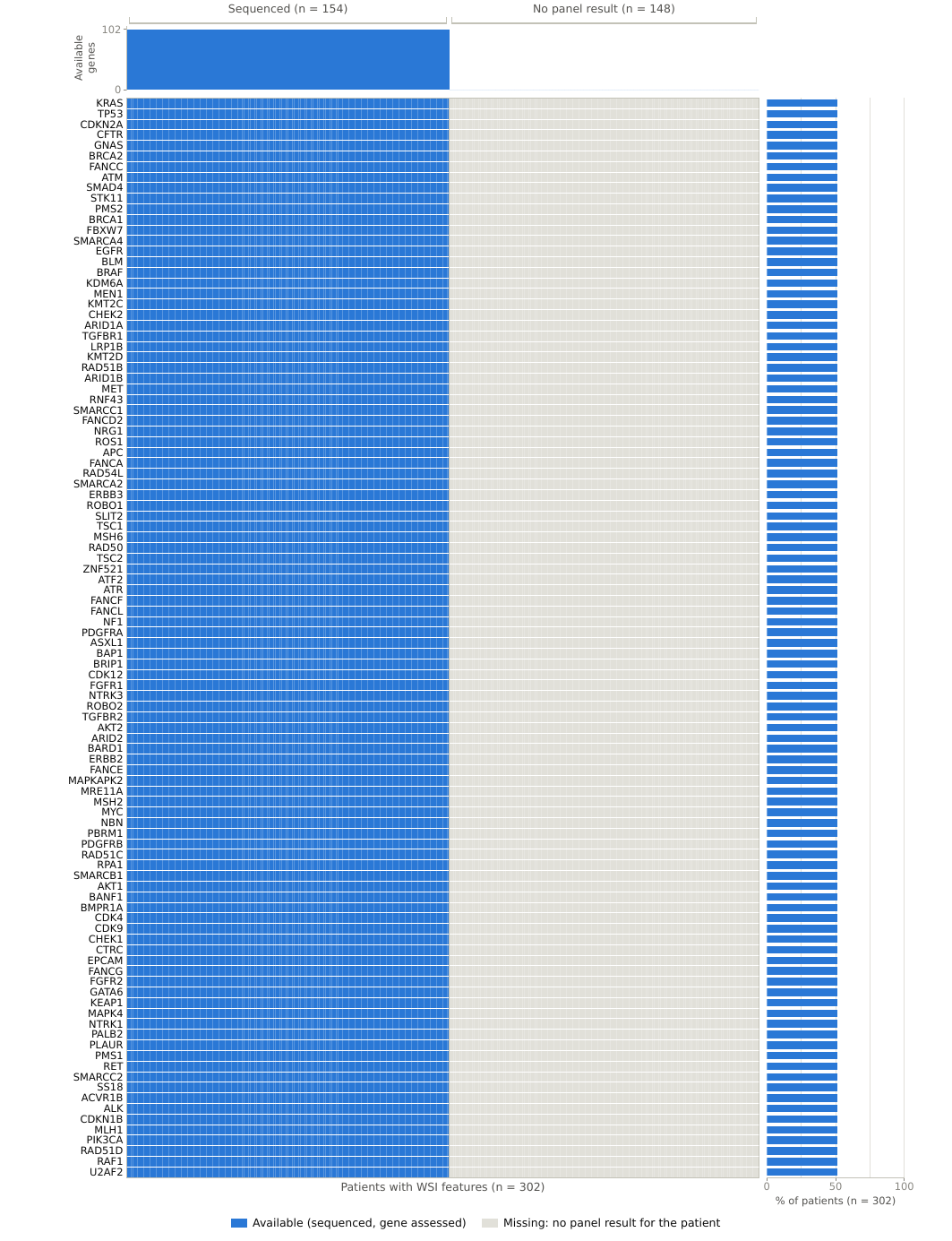}
    \caption{Availability of targeted sequencing data among 302 patients with whole-slide image features.}
    \label{fig:gene-availability}
\end{figure}

\subsection{Mutation profiles of sequenced patients}

The mutation matrix summarizes pathogenic or likely pathogenic variants across 102 genes in 154 patients with whole-slide image features. KRAS and TP53 were the most frequently mutated genes, affecting 133 (86.4\%) and 78 (50.6\%) patients, respectively. Mutations in other panel genes were uncommon. The number of mutated genes per patient ranged from zero to four.

\begin{figure}[t]
    \centering
    \includegraphics[width=\textwidth]{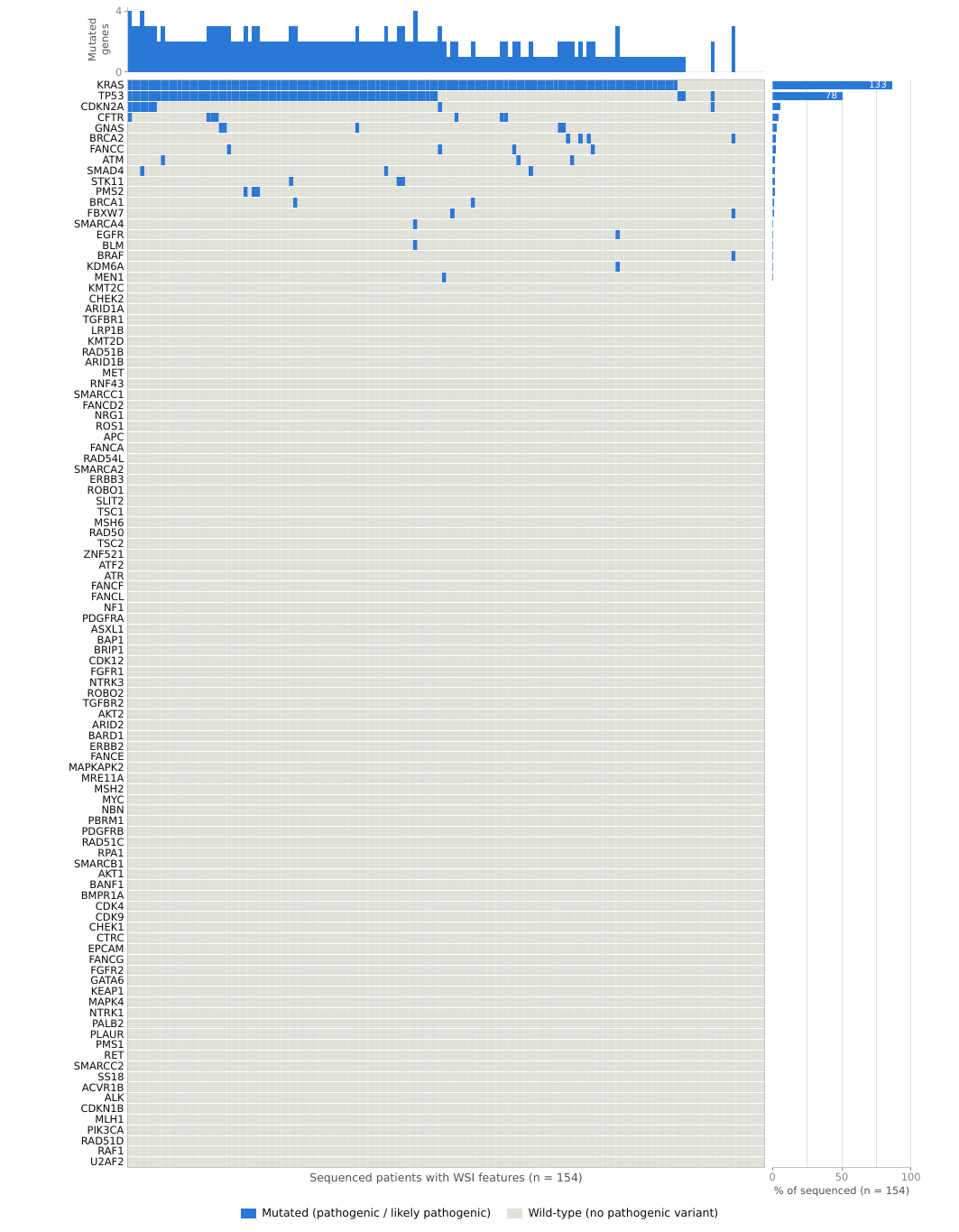}
    \caption{Mutational landscape of the sequenced patients of the WSI
    analysis cohort ($n = 154$).}
    \label{fig:mutations}
\end{figure}

\end{document}